\documentclass[lettersize,journal]{IEEEtran}
\usepackage{amsmath,amsfonts}
\usepackage{algorithmic}
\usepackage{array}
\usepackage{multirow}
\usepackage{booktabs}
\usepackage[caption=false,font=normalsize,labelfont=sf,textfont=sf]{subfig}
\usepackage{textcomp}
\usepackage{multirow}
\usepackage{float}
\usepackage{stfloats}
\usepackage{hyperref}

\hypersetup{
    colorlinks=true,
    linkcolor=blue,
    citecolor=blue,
    urlcolor=blue,
    }
\usepackage{makecell}
\usepackage{placeins}
\usepackage{caption}
\usepackage{url}
\usepackage{graphicx}
\def\BibTeX{{\rm B\kern-.05em{\sc i\kern-.025em b}\kern-.08em
    T\kern-.1667em\lower.7ex\hbox{E}\kern-.125emX}}
\usepackage{balance}

\begin{document}
\title{Supervised Learning and Large Language Model Benchmarks on Mental Health Datasets: Cognitive Distortions and Suicidal Risks in Chinese Social Media}
\author{Hongzhi Qi, Qing Zhao, Jianqiang Li, Changwei Song, Wei Zhai, Dan Luo, Shuo Liu, Yi Jing Yu, Fan Wang, Huijing Zou, Bing Xiang Yang, Guanghui Fu\textsuperscript{*}

\thanks{This work was supported by grants from the National Natural Science Foundation of China (grant numbers:72174152, 72304212 and 82071546), Fundamental Research Funds for the Central Universities (grant numbers: 2042022kf1218; 2042022kf1037), the Young Top-notch Talent Cultivation Program of Hubei Province.}

\thanks{Hongzhi Qi, Qing Zhao, Jianqiang Li, Changwei Song, Wei Zhai are with School of Software Engineering, Beijing University of Technology, Beijing, China.}
\thanks{Dan Luo, Shuo Liu, Yi Jing Yu, Fan Wang, Huijing Zou, Bing Xiang Yang are with School of Nursing, Wuhan University, Wuhan, China.}
\thanks{Guanghui Fu is with Sorbonne Universit\'{e}, Institut du Cerveau – Paris Brain Institute - ICM, CNRS, Inria, Inserm, AP-HP, H\^{o}pital de la Piti\'{e}-Salp\^{e}tri\`{e}re, F-75013, Paris, France.}
\thanks{Guanghui Fu is supported by a Chinese Government Scholarship provided by the China Scholarship Council (CSC).}
\thanks{\textsuperscript{*}Corresponding author: Guanghui Fu (\url{guanghui.fu@inria.fr})}
}

\markboth{Preprint}%
{Qi \MakeLowercase{\textit{et al.}}: }

\maketitle

\begin{abstract}
On social media, users often express their personal feelings, which may exhibit cognitive distortions or even suicidal tendencies on certain specific topics. Early recognition of these signs is critical for effective psychological intervention. In this paper, we introduce two novel datasets from Chinese social media: SOS-HL-1K for suicidal risk classification and SocialCD-3K for cognitive distortions detection. The SOS-HL-1K dataset contained 1,249 posts and SocialCD-3K dataset was a multi-label classification dataset that containing 3,407 posts. We propose a comprehensive evaluation using two supervised learning methods and eight large language models (LLMs) on the proposed datasets. From the prompt engineering perspective, we experimented with two types of prompt strategies, including four zero-shot and five few-shot strategies. We also evaluated the performance of the LLMs after fine-tuning on the proposed tasks. The experimental results show that there is still a huge gap between LLMs relying only on prompt engineering and supervised learning. In the suicide classification task, this gap is 6.95\% points in F1-score, while in the cognitive distortion task, the gap is even more pronounced, reaching 31.53\% points in F1-score. However, after fine-tuning, this difference is significantly reduced. In the suicide  and cognitive distortion classification tasks, the gap decreases to 4.31\% and 3.14\%, respectively. This research highlights the potential of LLMs in psychological contexts, but supervised learning remains necessary for more challenging tasks. All datasets and code are made available at: \url{https://github.com/HongzhiQ/SupervisedVsLLM-EfficacyEval}.
\end{abstract}


\section{Introduction} \label{sec:intro}

\IEEEPARstart{T}{he} prevalence of mental illness, particularly depression, continues to present significant challenges worldwide. According to the World Health Organization (WHO), an estimated 3.8\% of the global population experiences depression~\cite{who2023depressive}. Specifically in China, the prevalence of depression is notably high, with estimates around 6.9\%~\cite{huang2019prevalence}, underscoring the escalating mental health concerns in the nation. Such severe depression can often precipitate suicidal behaviors~\cite{who2023suicide}. Platforms such as Twitter and Sina Weibo have become channels for emotional outbursts as social media has become more widespread~\cite{keles2020systematic}. Within these platforms, a certain subset of topics recur, with users frequently expressing deep-seated negative emotions and even showing suicidal tendencies~\cite{robinson2016social, luxton2012social}.

Artificial intelligence (AI), especially the branches of deep learning and natural language processing, is a promising solution to this challenge~\cite{coppersmith2018natural}. Over recent years, AI research has resulted in the formulation of several algorithms tailored for emotion recognition within textual data~\cite{nandwani2021review}. However, these advancements are not without obstacles~\cite{acheampong2020text}. Constructing a potent deep learning model often demands considerable time and financial resources. The cost of data labeling, the need to engage domain experts, and the variability of model performance when moved to different application domains present significant challenges~\cite{saunders2022domain}. This highlights a critical need for more flexible and adaptable algorithmic solutions, especially in the mental health domain~\cite{laparra2020rethinking}. In this context, the emergence of large language models (LLMs) is particularly noteworthy, and they have already attracted significant attention in the mental health domain~\cite{he2023towards}.

Large language models, characterized by their large parameter and training datasets, stand as the state-of-the-art in the framework of computational linguistics~\cite{zhao2023survey, thirunavukarasu2023large}. Their potential lies in their ability to comprehend and emulate human-like text nuances. Despite their promising potential, several studies have attempted to validate their practical performance. For instance, Xu et al.~\cite{xu2023leveraging} examined four public datasets related to online social media sentiment detection. However, their study focused only on English data, and the classification task was relatively broad. To date, there is a notable gap in research concerning the Chinese context, particularly in the area of fine-grained emotion recognition, which is often of greater significance. For instance, the nuanced understanding of a wide spectrum of emotions can facilitate the early identification of mental health issues, allowing for timely and accurate mental health interventions. Such capabilities further highlight the importance of fine-grained emotion recognition in psychological assessments and therapy. Due to the lack of comprehensive evaluation and real-world testing, there are concerns about their use, especially in areas where high reliability is required, such as medical and healthcare~\cite{thirunavukarasu2023large}. 

However, so far, there is a lack of large-scale datasets for psychology-related tasks, especially cognitive distortions. 
Sharma et al.~\cite{sharma-etal-2023-cognitive} proposed an English dataset consists of 600 entries with 15 labels, 14 of which pertain to cognitive distortions, making it a multi-label classification dataset. Despite the diversity of cognitive distortion labels, the dataset's sample size is relatively small. The dataset, named Therapist Q\&A~\footnote{\url{https://www.kaggle.com/arnmaud/therapist-qa}} includes 2,507 entries with 11 labels, 10 of which are cognitive distortion labels, also categorized as a multi-label classification dataset.
To the best of our knowledge, there is only one Chinese cognitive distortions dataset, C2D2~\cite{wang2023c2d2}. However, they abandoned the multi-label annotation approach to save costs. In practice, it is common for a single text to reflect multiple cognitive distortions. 
For the suicide detection task, there is a lack of open datasets, particularly in Chinese. For instance, Cao et al.~\cite{cao-etal-2019-latent}  created a binary classification dataset for detecting suicide risk on Sina Weibo, the dataset includes 3,652 users identified as at risk of suicide and their 252,901 posts, as well as 3,677 users identified as not at risk and their 491,130 posts. However, it is important to note that not all of these posts explicitly express suicidal intent. Huang et al.~\cite{huang2017exploring} gathered account data from 130 Chinese Weibo users who were known to have died by suicide between 2011 and 2016. However, the provided link is not accessible.

To provide more contribution to the research community, we present two datasets: SOS-HL-1K for high and low suicide risk classification, and SocialCD-3K for a cognitive distortion task. 
the SOS-HL-1K dataset contains 1,249 social media posts reflect high and low suicide risk, and the SocialCD-3K dataset contains 3,407 social media posts annotated in a multi-label way. Both datasets were annotated by multiple domain experts. 
LLMs are prompt-driven and characterized by their flexibility and ease of use across different tasks. However, their specific performance in psychological tasks, especially in the Chinese domain, has not to be fully explored. To address this, we evaluate the performance of eight LLMs and test the performance of four zero-shot and five few-shot prompt strategies on the two proposed datasets. We also test the performance of different types of LLMs fine-tuned for these tasks. The experimental results show that LLMs based on prompt strategies perform well in the suicide risk classification task but poorly in the more challenging cognitive distortion classification task, lagging behind supervised learning by 6.95\% and 31.53\%, respectively. However, fine-tuned LLMs significantly improve their performance. While they still lag behind supervised learning, the gaps are narrowed to 4.31\% and 3.14\%, respectively. To the best of our knowledge, our proposed dataset is the first Chinese cognitive distortion multi-label classification dataset. We evaluated the performance of LLMs from different angles, providing a valuable reference for researchers using LLMs in psychological tasks.

\section{Related work} \label{sec:related} 

\subsection{Text sentiment analysis}\label{sec:realted:text}

Social media platforms have become central channels for the expression of emotions worldwide. Accurately and quickly identifying the emotions embedded in this data presents a big challenge to computational algorithms~\cite{nandwani2021review}. Fu et al.~\cite{fu2021distant} developed a distant supervision approach for classifying high and low suicide risk on Chinese social media platforms, using annotations by experts at different levels, significantly reducing the dependency on extensive manual effort by professional experts. Employing this approach, coupled with the integration of key psychological features extracted from user blogs, they achieved an F1 score of 77.98\%. Singh et al.~\cite{singh2021sentiment} focused on COVID-19 related sentiment analysis on social media platforms, training a BERT-based model on these two datasets, achieving an accuracy of 94\%. Wan~\cite{wan2019sentiment} introduced a method for sentiment analysis of Sina Weibo comments, leveraging deep neural networks. The data undergo feature extraction through multi-level pooling and convolution layers, and key features are subsequently extracted from the feature matrix using a CNN. For the final classification and sentiment analysis, the softmax logistic regression method is employed. Zhang et al.~\cite{zhang2020emotion} introduced a factor graph-based emotion recognition model that integrates correlations among emotion labels, social interactions, and temporal patterns into a unified framework, adeptly identifying multiple emotions through a multi-label learning approach.

Although deep learning algorithms demonstrate impressive performance, the requirement for significant volume of labeled data makes them difficult to apply. 
The utilization of distant supervision approach proposed by Fu et al.~\cite{fu2021distant} reduces the need for labeling but still involves expert groups. Given these limitations, there is a growing demand for efficient methods for emotion detection on social media. Recent advancements in LLMs offer potential solutions, attracting increased attention in the mental health domain.

\subsection{Large language model and its applications in medical domain}\label{sec:realted:llm}

The advent of Large Language Models (LLMs) like OpenAI's ChatGPT~\cite{zhao2023survey} has revolutionized natural language processing~\cite{kaddour2023challenges}, significantly outperforming smaller models~\cite{wei2022emergent}. These models have diverse applications, including content generation~\cite{liebrenz2023generating}, medical report assistance~\cite{jeblick2022chatgpt}, coding assistance~\cite{surameery2023use}, education~\cite{kasneci2023chatgpt}, and answering medical-related questions~\cite{yeo2023assessing}. Their large size of model enables them to generate complex, contextually relevant content, and make them flexible for use in downstream task.

LLMs have garnered significant attention in the medical domain~\cite{thirunavukarasu2023large}. Jiang et al.~\cite{jiang2023health} developed the clinical LLM NYUTron to assist physicians, achieving AUC scores between 78.7–94.9\%. It has been deployed in a prospective trial, indicating its potential for real-world application.
In psychology, Qin et al.~\cite{qin2023read} devised a depression detection system using LLMs, which raises ethical concerns. Chen et al.~\cite{chen2023llm} improved psychiatrist-patient simulations with ChatGPT-based chatbots, although limitations in realistic simulations were acknowledged. Fu et al.~\cite{fu2023enhancing} developed a counseling support system for non-professional counselors, achieving a 78\% expert approval rate. Ayers et al.~\cite{ayers2023comparing} found that ChatGPT-based chatbot responses were preferred over physicians' in online forums, though they recognized limitations in real-world applications.
Xu et al.~\cite{xu2023leveraging} evaluated multiple LLMs on mental health prediction tasks using English datasets, lacking multi-label classification tasks. Yang et al.~\cite{yang2023interpretable} assessed ChatGPT's capabilities in mental health analysis, finding it surpasses traditional methods but lags behind advanced, task-specific approaches.

In summary, while LLMs show promise in psychology and medicine, the performance in real-life situation requires further validation.

\section{Methods} \label{sec:methods}

\subsection{Baseline supervised learning models}\label{sec:method:baseline}

\begin{itemize}
    \item LSAN~\cite{xiao2019label}: The LSAN model uses label semantics to identify relationships between labels and documents, creating a label-specific document representation. It employs a self-attention mechanism to focus on this representation and uses an adaptive fusion strategy for multi-label text classification, proving effective in predicting low-frequency labels.

    \item BERT~\cite{devlin2018bert}: BERT uses a bidirectional approach, facilitated by the Transformer architecture~\cite{vaswani2017attention}, to understand the full context of words. It is pre-trained with a masked language model objective, predicting masked words. BERT excels in various NLP tasks, such as question-answering and sentiment analysis, especially when fine-tuned on specific data.
 
\end{itemize}

\subsection{Large language models}

For the LLM comparison experiments, we evaluated eight popular models, including two general LLMs: GPT-3.5, GPT-4~\cite{openai2023gpt4} and six Chinese-adaptive LLMs: ChatGLM2-6B, GLM-130B~\cite{zeng2022glm}, GLM-4~\cite{zhipu2024}, Chinese-Alpaca-2-7B~\cite{Chinese-LLaMA-Alpaca}, Chinese-LLaMa-2-7B~\cite{Chinese-LLaMA-Alpaca}, and Llama2-Chinese-7b-Chat~\cite{Llama2-Chinese-7b-Chat}.

\subsubsection{General LLMs: GPTs}
\begin{itemize}
    \item GPT-3.5: GPT-3.5 is an advanced iteration of the GPT-3~\cite{brown2020language} language model, offering improvements in conversational capabilities. It is designed to provide more coherent and context-aware responses in conversational applications, reflecting ongoing developments in language modeling techniques. 
    \item GPT-4~\cite{openai2023gpt4}: GPT-4 is a groundbreaking multimodal model that processes both images and text to generate text outputs. It performs at a human level on various benchmarks, including scoring in the top 10\% on a simulated bar exam. Built on the Transformer architecture~\cite{vaswani2017attention}, GPT-4 is trained to predict tokens and undergoes post-training alignment for improved accuracy.  Despite its capabilities, GPT-4 has limitations like occasional content hallucinations and a constrained context window.
\end{itemize}

\subsubsection{Chinese LLMs}
\begin{itemize}
    \item ChatGLM2-6B: ChatGLM2-6B is an open-source bilingual model with 6.2 billion parameters, optimized for Chinese question-answering and dialogue. Trained on about 1TB of Chinese and English text, it can be fine-tuned through various techniques like supervised learning and human feedback, allowing for diverse language processing applications.

    \item GLM-130B~\cite{zeng2022glm}: GLM-130B is a bilingual model with 130 billion parameters, optimized for English and Chinese. It aims to provide an open-source alternative comparable to GPT-3, outperforming GPT-3 175B on multiple English benchmarks and surpassing ERNIE TITAN 3.0 260B~\cite{wang2021ernie} on Chinese benchmarks.

    \item GLM-4~\cite{zhipu2024}: GLM-4 uses a bidirectional Transformer~\cite{vaswani2017attention} structure with the Masked and Prefix Language Model (MPCT) pre-training strategy, combining MLM and PLM strengths. It employs Rotary Position Embedding (RoPE) for long text sequences, adaptive masking for varied tasks, and supports multi-task learning.

    \item Chinese-LLaMa-2-7B~\cite{Chinese-LLaMA-Alpaca}: Chinese-LLaMa-2-7B, based on LLaMa-2~\cite{touvron2023llama}, uses a new vocabulary of 55,296 entries to enhance Chinese text coverage. It employs Low-Rank Adaptation (LoRA)~\cite{hu2021lora} for fine-tuning, supporting multi-task learning, and demonstrating superior performance in Chinese tasks.

    \item Chinese-Alpaca-2-7B~\cite{Chinese-LLaMA-Alpaca}: Chinese-Alpaca-2-7B focuses on conversational and instruction-following capabilities. Using LoRA~\cite{hu2021lora} for efficient training, it excels in multi-turn dialogues and complex command parsing after supervised fine-tuning.

    \item Llama2-Chinese-7b-Chat~\cite{Llama2-Chinese-7b-Chat}: Built on LLaMa-2~\cite{touvron2023llama}, Llama2-Chinese-7b-Chat is optimized for Chinese through continuous pre-training on a large-scale dataset. It leverages LoRA~\cite{hu2021lora} for training, enhancing conversational capabilities with Chinese-specific commands and data.

\end{itemize}


\subsection{LLM prompt strategies}

LLMs use prompts as guides and are valued for their flexibility in performing downstream tasks. However, it is crucial to design the prompts precisely, as this directly affects the LLM's understanding of the given task and the output of the results~\cite{lin2024write}. To this end, we design the following prompt strategies, including zero-shot and few-shot approaches.
Zero-shot prompts denote direct interaction with the LLM without providing any task examples, while few-shot prompts involve interaction with the LLM by giving some examples within tasks, similar to the training data used in the supervised learning.

\subsubsection{Zero-shot prompting} \label{sec:method:llm_zero}

We begin our experiment with a prompt design tailored to tasks within a zero-shot paradigm, and it includes several strategies: direct task requests (represent as basic), role definition, scene definition, and a hybrid strategy. For illustrative purposes, the cognitive distortion classification task is used as an example. Note that, since our experimental data are in Chinese, we also use Chinese prompts. We provide translations of the English prompts in the following explanation. To avoid repetitions in prompts, we selected the key part of each strategy and represented it as $\texttt{Strategy}$ [Prompt]. When the prompt was a combination of several strategies from $1$ to $n$, it is referred to as $\texttt{Strategy}_1+\texttt{Strategy}_2+\cdots+\texttt{Strategy}_n$.

\begin{enumerate}
    \item Basic: A direct task statement without specific contextual emphasis. Prompt English translation: ``$\texttt{Basic}$ [Please perform a multi-classification task to determine whether it reflects any of the specified 12 cognitive distortions: (list of cognitive distortions categories)].''

    \item Role-definition prompting: This strategy delineates the role of the LLM (in this case, a psychologist) and emphasizes the need for psychological expertise. Prompt English translation: ``$\texttt{Role}$ [Assuming the role of a psychologist with professional psychological experiences] $+\texttt{Basic}$''

    \item Scene-definition prompting: Introduces the context of a social media environment, highlighting user identifiers to eliminate ambiguity. Prompt English translation: ``$\texttt{Scene}$ [Given the user ID $u$ and the associated posts on social media, based on the post content], $+ \texttt{Basic}$''

    \item Hybrid prompting: A combination of both role and scene definitions that provide integrated prompt, and it can be represent as $\texttt{Scene}+\texttt{Role}+\texttt{Basic}$.

\end{enumerate}

\subsubsection{Few-shot prompting} \label{sec:method:llm_few}
Few-shot prompting is construed as a method to integrate prior knowledge or a batch of $n$ training samples into LLMs, thereby enabling them to learn this information and adeptly execute the task. We experiment with the following setup.

\begin{enumerate}
    \item Background knowledge: This strategy is provided with psychological definitions, supplemented by representative cases, followed by one of the four prompting strategies devised from zero-shot prompting. Prompts that integrate background knowledge and employ the hybrid strategy from zero-shot prompting are detailed as follows: ``$\texttt{Background}$ [There are the definitions of 12 cognitive distortions (list of cognitive distortions definitions), and there are the representative cases (list of cognitive distortions cases). Consider these cognitive distortions definitions and cases], $+\texttt{Scene}+\texttt{Role}+\texttt{Basic}$.''

    \item Prompting with $n$ reference samples per category: In this strategy, reference data ${\mathrm{ref}}_{n}$ are randomly selected from training set for each category to conduct the prompt of LLM, followed by one of the four prompting strategies designed from zero-shot prompting. Prompts that incorporate the reference data and employ the hybrid strategy from zero-shot prompting are detailed as: ``$\texttt{Reference}$ [There are some examples of the target task with the ground truth label (list of reference samples). Consider these cognitive distortions examples], $+\texttt{Scene}+\texttt{Role}+\texttt{Basic}$.''

   \item Background knowledge and prompting with $n$ reference samples per category: This approach investigates whether enhancing explanations in few-shot prompting can improve the LLM’s understanding of psychological tasks. It incorporates psychological definitions, and provides $n$  reference samples per category for LLM prompt conduction. The following example integrates background knowledge and reference instances, and employ the hybrid strategy from zero-shot prompting. It can be represented as $\texttt{Background}+\texttt{Reference}+\texttt{Scene}+\texttt{Role}+\texttt{Basic}$.
\end{enumerate}

\subsection{LLM fine-tuning for downstream task}

Fine-tuning LLM is the process of adapting a pre-trained language model to a specific task or domain by continuing its training on a smaller, task-specific dataset~\cite{ding2023parameter}. Fine-tuning typically involves updating the model's weights using supervised learning with labeled data, enhancing its performance in the targeted downstream task. Following the fine-tuning, our evaluation paradigm retained the role, scene and hybrid definitions from the zero-shot prompting for consistency and comparative assessment as introduced in Section~\ref{sec:method:llm_zero}.

\section{Experiments and results} \label{sec:experiments}

\subsection{Datasets and evaluation metrics}
We proposed two psychology-related classification datasets for high/low suicide risk classification and cognitive distortion multi-label classification, named ``SOS-HL-1K'' and ``SocialCD-3K''. The suicide risk task primarily differentiates between high and low suicide risks, while the cognitive distortion task focuses on classifications defined by Burns~\cite{burns1981feeling}. 
We sourced our data by crawling comments from the ``Zoufan'' blog within the Weibo social platform. Subsequently, a team of qualified psychologists were enlisted to annotate the data. In our study, strict data preprocessing measures were adopted to protect users’ privacy.

The suicide detection dataset is a binary classification dataset consisting of 1249 samples. Among these, there are 648 posts with low suicide risk and 601 posts with high suicide risk. The cognitive distortion dataset contains a total of 3407 posts. This is a multi-label classification dataset, with each sample corresponding to one or more cognitive distortion categories. The classification labels used for this data follow the definitions provided by Burns~\cite{burns1981feeling}.
For both sets of data, the training set and test set are divided according to the ratio of 4:1. 
The data distribution and specific categories of these two datasets are listed in Table~\ref{tab:Datasets}.
We utilize three evaluation metrics to measure the performance of different algorithms for our two tasks: precision, recall, and F1-score. 
For the multi-label classification task of cognitive distortions, we report the micro-averaged F1-score. This method aggregates the contributions of all classes to compute the average score, thereby offering a more comprehensive view of the model's performance, particularly useful in scenarios with unbalanced distribution of class instances.

\begin{table}
\centering
\caption{Data distribution of the proposed datasets: SOS-HL-1K and SocialCD-3K for high/low suicide risk and cognitive distortion classification on social media. $N_{c}$ indicate the number of posts for the corrsponding categories. $N_{train}$ and $N_{test}$ represent the number of training and test set, respectively. $L$ is the total number of classes, $\overline{L}$ is the average number of labels per sample, and $\overline{C}$  is the average number of words per post.}
\label{tab:Datasets}
\begin{tabular}{|c|l|c|} 
\hline
Dataset                       & \multicolumn{1}{c|}{Categories} & $N_{c}$                                                              \\ 
\hline
\multirow{3}{*}{SOS-HL-1K}    & High risk                     & 601                                                                     \\
                              & Low risk                      & 648                                                                     \\ 
\cline{2-3}
                              & \multicolumn{2}{l|}{$N_{train}$=999, $N_{test}$=250, $L$=1, $\overline L$=1, $\overline C$=47.79}       \\
\hline
\multirow{13}{*}{SocialCD-3K} & All-or-nothing thinking       & 77                                                                      \\
                              & Over-generalization           & 141                                                                     \\
                              & Mental filter                 & 378                                                                     \\
                              & Disqualifying the positive    & 27                                                                      \\
                              & Mind reading                  & 121                                                                     \\
                              & The fortune teller error      & 652                                                                     \\
                              & Magnification                 & 321                                                                     \\
                              & Emotional reasoning           & 16                                                                      \\
                              & Should statements             & 84                                                                      \\
                              & Labeling and mislabeling      & 1961                                                                    \\
                              & Blaming oneself~              & 188                                                                     \\
                              & Blaming others                & 27                                                                      \\ 
\cline{2-3}
                              & \multicolumn{2}{l|}{$N_{train}$=2725, $N_{test}$=682, $L$=12, $\overline L$=1.71, $\overline C$=42.56}  \\ 
\hline

\end{tabular}
\end{table}

\subsection{Experiment design}

Our experiments focus on validating the two proposed tasks, particularly testing different strategies for using LLMs. This includes four zero-shot and three few-shot prompt construction strategies, as well as the performance of fine-tuned LLMs. 
In total, we compare two types of supervised learning methods and eight LLMs. Using cognitive distortions as an example to show our points, our evaluations spanned several dimensions:
\begin{itemize}
    \item Prompt conduction: Initially, we assessed four prompting strategies within the zero-shot learning conduction. Subsequently, based on their performance, the top two strategies were selected for further evaluation in the few-shot learning setting across various LLMs.
    \item LLM ability: In our experiments, we found that the performance of the ChatGLM2-6B and GLM-130B models was completely fail. Therefore, we decided to use the latest version of the GLM series, the GLM-4 model, to evaluate performance on the newly expanded dataset. For GPT-3.5, its token limitation prevented us from entering five samples for each category during few-shot prompting. Consequently, we reserved the ${\mathrm{ref}}_{5}$ approach exclusively for GPT-4. 

    \item LLM fine-tuning: OpenAI recently introduced a fine-tuning feature for GPT-3.5, and official reports suggest that, under certain conditions, fine-tuned GPT-3.5 may outperform GPT-4~\cite{gpt35finetune}. Therefore, we experimented with fine-tuning GPT-3.5. Since the current version of GPT-4 lacks fine-tuning capabilities, we were unable to assess its potential. Additionally, we explored the fine-tuning performance of three other open-source Chinese large models in the cognitive distortion recognition task.
\end{itemize}
Note that, since the experimental performance of the three Chinese LLMs (Chinese-Alpaca-2-7B, Chinese-LLaMa-2-7B, and Llama2-Chinese-7b-Chat) on cognitive distortion was similar to the performance of GPT-3.5 after fine-tuning, we did not conduct experiments with these three LLMs in the suicide risk classification task.
Also, ChatGLM2-6B's performance in the cognitive distortion classification task was too poor to even output in the correct format, so we did not include its experimental results. By the time we experimented with the cognitive distortion task, GLM-130B had been replaced by a newer version, GLM-4, so we evaluated GLM-4 in this task instead.

\subsection{Implementation details}

\begin{itemize}
    \item LSAN: We used word2vec~\cite{mikolov2013efficient} to train 300-dimensional embeddings for both document and randomly-initialized label texts. The function of attention mechanism is to compute word contributions to labels and create label-specific document representations. Dot products between these document and label vectors refined these relationships further. These two types of document representations were then fused using weighted combinations. For predictions, we employed a fully connected layer, followed by RELU and a sigmoid function. We used cross-entropy as loss function.
    
    \item BERT: We employ Chinese pretrained BERT to extract 768-dimensional vectors from the sentences. To avoid overfitting, a dropout function~\cite{srivastava2014dropout} is applied to these sentence vectors. Subsequently, a fully connected layer is introduced for classification. The sigmoid function serves as the activation function for the output layer.

    \item LLM-zero shot: Both GPT-3.5 and GPT-4~\cite{openai2023gpt4} are closed-source and available through OpenAI's API. We used gpt-3.5-turbo for it is one of the most capable and cost-effective models in the GPT-3.5 family, and GPT-4 for its advanced capabilities. For GLM models, we deployed the open-source ChatGLM2-6B~\cite{zeng2022glm} on our server, and tested the larger GLM-130B~\cite{zeng2022glm} via its official website due to deployment challenges. We averaged the performance over five rounds of experiments for all models to minimize randomness. For GPT-3.5, GPT-4, ChatGLM2-6B and GLM-4 we set temperatures to 0.1, 0.3, 0.5, 0.7, and 0.9, and averaged the results over five rounds of experiments. For GLM-130B, we could not adjust the temperature due to its limitations.

    \item LLM-few shot: We conducted the experiments using the top two performing prompt strategies from the zero-shot performance, as determined by their F1-scores. 
    Given the different input token constraints of each model, we selected varying amounts of reference data $\mathrm{ref}_n$ according to the requirements of the corresponding models.

    \item LLM-fine-tunning: For the closed-source model GPT-3.5, we used the API provided by OpenAI to fine-tune the GPT-3.5 Turbo model for suicide risk and cognitive distortions tasks. In this experiment, the training epoch was set to 3. For other hyperparameters, we did not explicitly specify them; instead, OpenAI selected the default values based on the dataset size. We also explored the performance of three open-source Chinese LLMs (Chinese-LLaMa-2-7B, Chinese-Alpaca-2-7B~\cite{Chinese-LLaMA-Alpaca}, Llama2-Chinese-7b-Chat~\cite{Llama2-Chinese-7b-Chat}) on this fine-tuning task. The open-source models were deployed locally for experiments. For these three models, we were deployed on an NVIDIA A100 GPU, and all were fine-tuned using the LoRA~\cite{hu2021lora}. The batch size was set to 8, save steps were set to 300, the learning rate was set to 6e-5, the learning rate scheduler was cosine, and the number of epochs was set to 5.
\end{itemize}

\section{Results}
In our study, we primarily aim to answer two questions: how can LLMs be most simply used to accomplish a task, and what is the performance potential of LLMs on both tasks. The order of using LLMs ranges from simple to complex, starting with zero-shot and few-shot prompt engineering to the fine-tuning. This analysis is detailed in the following sections, and the results of suicide classification and cognitive distortions classification can be seen in Table~\ref{tab:results_suicide} and Table~\ref{tab:results_cognitive_new} respectively.

\begin{table*}[htbp]
\centering
\caption{Result for suicide binary classification task. All these methods were evaluated on the same test set contains 250 posts. The terms ``zero-shot'' and ``few-shot'' refer to different prompt construction without training. `Train from scratch` and ``fine-tuning'' indicates that the corresponding method was trained or fine-tuned using the training set.}
\label{tab:results_suicide}
\resizebox{0.9\linewidth}{!}{
\begin{tabular}{|c|c|l|l|c|c|c|c|} 
\hline
Model category                       & Model                    & Type                         & Sub-type                          & Train data & Precision & Recall  & F1-score  \\ 
\hline
\multirow{2}{*}{Supervised learning} & LSAN                          & Train from scarch            & -                                 & 999        & 74.59\%   & 87.50\% & 80.53\%   \\
                                     & BERT                          & Fine-tuning                  & -                                 & 999        & 88.42\%   & 77.78\% & 82.76\%   \\ 
\hline
\multirow{51}{*}{LLM}                & \multirow{14}{*}{ChatGLM2-6B} & \multirow{4}{*}{Zero-shot}   & basic                             & 0          & 69.07\%   & 37.10\% & 48.07\%   \\
                                     &                               &                              & role                       & 0          & 65.77\%   & 35.81\% & 46.15\%   \\
                                     &                               &                              & scene                      & 0          & 64.52\%   & 45.16\% & 53.01\%   \\
                                     &                               &                              & hybrid                            & 0          & 65.68\%   & 46.13\% & 53.74\%   \\ 
\cline{3-8}
                                     &                               & \multirow{10}{*}{Few-shot}   & background+scene           & 0          & 58.56\%   & 47.26\% & 51.45\%   \\
                                     &                               &                              & background+hybrid                 & 0          & 60.37\%   & 70.64\% & 64.41\%   \\
                                     &                               &                              & ${\mathrm{ref}}_{12}$+scene            & 24         & 67.19\%   & 59.52\% & 63.04\%   \\
                                     &                               &                              & ${\mathrm{ref}}_{12}$+hybrid                  & 24         & 64.29\%   & 49.20\% & 55.56\%   \\
                                     &                               &                              & background+${\mathrm{ref}}_{12}$+scene & 24         & 49.74\%   & 56.61\% & 52.70\%   \\
                                     &                               &                              & background+${\mathrm{ref}}_{12}$+hybrid       & 24         & 58.91\%   & 73.23\% & 64.78\%   \\
                                     &                               &                              & ${\mathrm{ref}}_{30}$+scene            & 60         & 57.71\%   & 26.77\% & 36.14\%   \\
                                     &                               &                              & ${\mathrm{ref}}_{30}$+hybrid                  & 60         & 50.60\%   & 24.84\% & 32.60\%   \\
                                     &                               &                              & background+${\mathrm{ref}}_{30}$+scene & 60         & 62.97\%   & 52.90\% & 57.02\%   \\
                                     &                               &                              & background+${\mathrm{ref}}_{30}$+hybrid       & 60         & 60.14\%   & 47.10\% & 51.88\%   \\ 
\cline{2-8}
                                     & \multirow{10}{*}{GLM-130B}    & \multirow{4}{*}{Zero-shot}   & basic                             & 0          & 54.58\%   & 95.81\% & 69.52\%   \\
                                     &                               &                              & role                       & 0          & 55.51\%   & 94.84\% & 70.02\%   \\
                                     &                               &                              & scene                      & 0          & 55.05\%   & 93.87\% & 69.40\%   \\
                                     &                               &                              & hybrid                            & 0          & 57.37\%   & 97.42\% & 72.20\%   \\ 
\cline{3-8}
                                     &                               & \multirow{6}{*}{Few-shot}    & background+role            & 0          & 56.55\%   & 90.32\% & 69.55\%   \\
                                     &                               &                              & background+hybrid                 & 0          & 56.91\%   & 92.42\% & 70.43\%   \\
                                     &                               &                              & ${\mathrm{ref}}_{12}$+role             & 24         & 53.18\%   & 83.23\% & 64.89\%   \\
                                     &                               &                              & ${\mathrm{ref}}_{12}$+hybrid                  & 24         & 55.30\%   & 88.39\% & 68.02\%   \\
                                     &                               &                              & background+${\mathrm{ref}}_{12}$+role  & 24         & 57.84\%   & 83.38\% & 68.30\%   \\
                                     &                               &                              & background+${\mathrm{ref}}_{12}$+hybrid       & 24         & 60.88\%   & 90.00\% & 72.61\%   \\ 
\cline{2-8}
                                     & \multirow{13}{*}{GPT-3.5}     & \multirow{4}{*}{Zero-shot}   & basic                             & 0          & 52.00\%   & 88.23\% & 65.42\%   \\
                                     &                               &                              & role                       & 0          & 53.31\%   & 96.13\% & 68.59\%   \\
                                     &                               &                              & scene                      & 0          & 52.16\%   & 89.03\% & 65.76\%   \\
                                     &                               &                              & hybrid                            & 0          & 52.55\%   & 92.26\% & 66.95\%   \\ 
\cline{3-8}
                                     &                               & \multirow{6}{*}{Few-shot~}   & background+role            & 0          & 54.90\%   & 88.55\% & 67.76\%   \\
                                     &                               &                              & background+hybrid                 & 0          & 55.27\%   & 89.03\% & 68.19\%   \\
                                     &                               &                              & ${\mathrm{ref}}_{12}$+role             & 24         & 56.34\%   & 83.39\% & 67.22\%   \\
                                     &                               &                              & ${\mathrm{ref}}_{12}$+hybrid                  & 24         & 57.19\%   & 84.68\% & 68.27\%   \\
                                     &                               &                              & background+${\mathrm{ref}}_{12}$+role  & 24         & 59.37\%   & 81.61\% & 68.71\%   \\
                                     &                               &                              & background+${\mathrm{ref}}_{12}$+hybrid       & 24         & 58.26\%   & 82.90\% & 68.41\%   \\ 
\cline{3-8}
                                     &                               & \multirow{3}{*}{Fine-tuning} & role                       & 999        & 84.76\%   & 71.77\% & 77.73\%   \\
                                     &                               &                              & scene                      & 999        & 84.11\%   & 72.58\% & 77.92\%   \\
                                     &                               &                              & hybrid                            & 999        & 84.26\%   & 73.39\% & 78.45\%   \\ 
\cline{2-8}
                                     & \multirow{14}{*}{GPT-4}       & \multirow{4}{*}{Zero-shot}   & basic                             & 0          & 57.43\%   & 95.48\% & 71.72\%   \\
                                     &                               &                              & role                       & 0          & 57.29\%   & 97.26\% & 72.10\%   \\
                                     &                               &                              & scene                      & 0          & 58.81\%   & 97.58\% & 73.39\%   \\
                                     &                               &                              & hybrid                            & 0          & 57.47\%   & 97.42\% & 72.30\%   \\ 
\cline{3-8}
                                     &                               & \multirow{10}{*}{Few-shot}   & background+scene           & 0          & 64.91\%   & 73.55\% & 68.86\%   \\
                                     &                               &                              & background+hybrid                 & 0          & 63.24\%   & 84.03\% & 72.05\%   \\
                                     &                               &                              & ${\mathrm{ref}}_{12}$+scene            & 24         & 60.70\%   & 94.35\% & 73.87\%   \\
                                     &                               &                              & ${\mathrm{ref}}_{12}$+hybrid                  & 24         & 59.77\%   & 84.19\% & 69.87\%   \\
                                     &                               &                              & background+${\mathrm{ref}}_{12}$+scene & 24         & 65.44\%   & 81.77\% & 72.63\%   \\
                                     &                               &                              & background+${\mathrm{ref}}_{12}$+hybrid       & 24         & 65.65\%   & 78.87\% & 71.60\%   \\
                                     &                               &                              & ${\mathrm{ref}}_{30}$+scene            & 60         & 61.11\%   & 92.42\% & 73.56\%   \\
                                     &                               &                              & ${\mathrm{ref}}_{30}$+hybrid                  & 60         & 60.79\%   & 89.03\% & 72.22\%   \\
                                     &                               &                              & background+${\mathrm{ref}}_{30}$+scene & 60         & 63.86\%   & 83.06\% & 72.16\%   \\
                                     &                               &                              & background+${\mathrm{ref}}_{30}$+hybrid       & 60         & 70.16\%   & 82.58\% & 75.81\%   \\
\hline
\end{tabular}
}
\end{table*}

\begin{table*}[htbp]
\centering
\caption{Result for cognitive distortion multi-label classification task. All these methods were evaluated on the same test set contains 682 posts. The terms ``zero-shot'' and ``few-shot'' refer to different prompt construction without training. `Train from scratch` and ``fine-tuning'' indicates that the corresponding method was trained or fine-tuned using the training set.}
\label{tab:results_cognitive_new}
\resizebox{1\linewidth}{!}{
\begin{tabular}{|c|c|l|l|c|c|c|c|} 
\hline
Model category                       & Model                             & Type                         & Sub-type                        & Train data & Precision & Recall  & F1-score  \\ 
\hline
\multirow{2}{*}{Supervised learning} & LSAN                                    & Train from scratch           & -                               & 2725       & 75.53\%   & 71.48\% & 73.45\%   \\
                                     & BERT                                    & Fine-tuning                  & -                               & 2725       & 85.43\%   & 68.62\% & 76.10\%   \\ 
\hline
\multirow{44}{*}{LLM}                & \multirow{12}{*}{GLM-4}                 & \multirow{4}{*}{Zero-shot}   & basic                           & 0          & 17.39\%   & 46.95\% & 25.38\%   \\
                                     &                                         &                              & role                     & 0          & 20.56\%   & 46.95\% & 28.59\%   \\
                                     &                                         &                              & scene                    & 0          & 17.19\%   & 45.33\% & 24.93\%   \\
                                     &                                         &                              & hybrid                          & 0          & 19.54\%   & 47.32\% & 27.66\%   \\ 
\cline{3-8}
                                     &                                         & \multirow{8}{*}{Few-shot}    & background+hybrid               & 0          & 30.63\%   & 48.82\% & 37.64\%   \\
                                     &                                         &                              & background+role          & 0          & 28.76\%   & 47.20\% & 35.74\%   \\
                                     &                                         &                              & ${\mathrm{ref}}_{2}$+hybrid                 & 24         & 32.25\%   & 43.21\% & 36.94\%   \\
                                     &                                         &                              & ${\mathrm{ref}}_{2}$+role            & 24         & 34.82\%   & 44.58\% & 39.10\%   \\
                                     &                                         &                              & background+${\mathrm{ref}}_{2}$+hybrid      & 24         & 41.74\%   & 47.82\% & 44.57\%   \\
                                     &                                         &                              & background+${\mathrm{ref}}_{2}$+role & 24         & 41.91\%   & 45.83\% & 43.78\%   \\
                                     &                                         &                              & ${\mathrm{ref}}_{5}$+hybrid                 & 60         & 26.32\%   & 34.74\% & 29.95\%   \\
                                     &                                         &                              & ${\mathrm{ref}}_{5}$+role            & 60         & 29.67\%   & 36.61\% & 32.78\%   \\ 
\cline{2-8}
                                     & \multirow{11}{*}{GPT-3.5}               & \multirow{4}{*}{Zero-shot}   & basic                           & 0          & 10.63\%   & 13.95\% & 12.06\%   \\
                                     &                                         &                              & role                     & 0          & 12.17\%   & 10.21\% & 11.10\%   \\
                                     &                                         &                              & scene                    & 0          & 10.33\%   & 11.83\% & 11.03\%   \\
                                     &                                         &                              & hybrid                          & 0          & 10.59\%   & 11.83\% & 11.18\%   \\ 
\cline{3-8}
                                     &                                         & \multirow{4}{*}{Few-shot}    & background+hybrid               & 0          & 20.16\%   & 12.20\% & 15.21\%   \\
                                     &                                         &                              & background+basic                & 0          & 25.84\%   & 11.46\% & 15.88\%   \\
                                     &                                         &                              & ${\mathrm{ref}}_{2}$+hybrid                 & 24         & 16.76\%   & 19.68\% & 18.10\%   \\
                                     &                                         &                              & ${\mathrm{ref}}_{2}$+basic                  & 24         & 17.61\%   & 14.69\% & 16.02\%   \\ 
\cline{3-8}
                                     &                                         & \multirow{3}{*}{Fine-tuning} & scene                    & 2725       & 72.03\%   & 71.86\% & 71.95\%   \\
                                     &                                         &                              & role                     & 2725       & 69.88\%   & 70.49\% & 70.18\%   \\
                                     &                                         &                              & hybrid                          & 2725       & 71.94\%   & 72.48\% & 72.21\%   \\ 
\cline{2-8}
                                     & \multirow{3}{*}{Chinese-Alpaca-2-7B}    & \multirow{3}{*}{Fine-tuning} & scene                    & 2725       & 72.61\%   & 71.07\% & 71.83\%   \\
                                     &                                         &                              & role                     & 2725       & 73.60\%   & 72.32\% & 72.96\%   \\
                                     &                                         &                              & hybrid                          & 2725       & 72.02\%   & 69.95\% & 70.97\%   \\ 
\cline{2-8}
                                     & \multirow{3}{*}{Chinese-Llama-2-7B}     & \multirow{3}{*}{Fine-tuning} & scene                    & 2725       & 73.56\%   & 71.45\% & 72.49\%   \\
                                     &                                         &                              & role                     & 2725       & 72.82\%   & 68.83\% & 70.77\%   \\
                                     &                                         &                              & hybrid                          & 2725       & 73.46\%   & 70.07\% & 71.73\%   \\ 
\cline{2-8}
                                     & \multirow{3}{*}{Llama2-Chinese-7b-Chat} & \multirow{3}{*}{Fine-tuning} & scene                    & 2725       & 66.22\%   & 61.85\% & 63.96\%   \\
                                     &                                         &                              & role                     & 2725       & 68.59\%   & 65.09\% & 66.80\%   \\
                                     &                                         &                              & hybrid                          & 2725       & 69.44\%   & 64.59\% & 66.93\%   \\ 
\cline{2-8}
                                     & \multirow{12}{*}{GPT-4}                 & \multirow{4}{*}{Zero-shot}   & basic                           & 0          & 17.13\%   & 59.65\% & 26.61\%   \\
                                     &                                         &                              & role                     & 0          & 17.30\%   & 57.53\% & 26.61\%   \\
                                     &                                         &                              & scene                    & 0          & 18.29\%   & 42.71\% & 25.62\%   \\
                                     &                                         &                              & hybrid                          & 0          & 17.41\%   & 56.41\% & 26.61\%   \\ 
\cline{3-8}
                                     &                                         & \multirow{8}{*}{Few-shot}    & background+hybrid               & 0          & 30.68\%   & 53.67\% & 39.04\%   \\
                                     &                                         &                              & background+basic                & 0          & 31.38\%   & 56.66\% & 40.39\%   \\
                                     &                                         &                              & ${\mathrm{ref}}_{2}$+hybrid                 & 24         & 34.62\%   & 40.35\% & 37.26\%   \\
                                     &                                         &                              & ${\mathrm{ref}}_{2}$+basic                  & 24         & 37.56\%   & 40.97\% & 39.19\%   \\
                                     &                                         &                              & background+${\mathrm{ref}}_{2}$+hybrid      & 24         & 26.25\%   & 43.84\% & 32.84\%   \\
                                     &                                         &                              & background+${\mathrm{ref}}_{2}$+basic       & 24         & 37.56\%   & 40.97\% & 39.19\%   \\
                                     &                                         &                              & ${\mathrm{ref}}_{5}$+hybrid                 & 60         & 26.37\%   & 47.82\% & 34.00\%   \\
                                     &                                         &                              & ${\mathrm{ref}}_{5}$+basic                  & 60         & 29.77\%   & 53.80\% & 38.33\%   \\
\hline
\end{tabular}
}
\end{table*}

For the zero-shot prompting strategy, it can achieve strong performance for the suicide classification task, achieving an F1-score of around 70\% for all models except the smallest, ChatGLM2-6B. The size of model and limited number of pre-training data diminish ChatGLM2-6B's ability as an LLM, resulting in its failure in the cognitive distortion task. Conversely, the larger GLM-130B performs much better, surpassing ChatGLM2-6B by about 20\% points in F1-score and achieving similar results to GPT-3.5 and GPT-4. Nonetheless, GPT-4 generally outperforms the other LLMs.
Different models benefit from different prompt construction strategies. For instance, the hybrid strategy works best for GLM-130B, the role-define strategy is optimal for GPT-3.5, and the scene-define strategy yields the best results for GPT-4. However, prompt do not significantly affect performance. While the zero-shot strategy performs well overall in the suicide risk classification task, it is ineffective in the cognitive distortion recognition task. The best-performing zero-shot model, GLM-4 role-define, achieves an F1-score that is 47.51\% points lower than the supervised learning BERT, indicating that the zero-shot strategy, relying solely on prompt engineering, is adequate for simpler tasks like suicide risk classification but falls short for more complex tasks like cognitive distortion recognition.
Careful tuning of prompt can lead to performance gains, but the optimal strategy varies from model to model and task to task, necessitating a case-by-case approach. Generally, setting up the situation reasonably well tends to be beneficial.

The few-shot prompting strategy involves introducing part of the training data as reference to construct a prompt that guides the model in completing the task. For the suicide risk classification task, this strategy does not significantly improve on the zero-shot results. Interestingly, for ChatGLM2-6B, this strategy improves the F1-score by 11.08\% points, particularly under the ${\mathrm{ref}}_{12}$ related strategy. However, further increasing the data to ${\mathrm{ref}}_{30}$ does not enhance performance; instead, it decreases.
For other models, the few-shot strategy yields almost no improvement in the suicide classification task. Only the GPT-4 model shows a gain of 3.51\% points in F1-score using the background+${\mathrm{ref}}_{30}$+hybrid strategy. In the cognitive distortion task, all models continue to perform poorly under this strategy, with most achieving around a 40\% F1-score. However, it is noteworthy that this strategy improves upon the zero-shot approach, with F1-score gains of 16.91\% points for the GLM-4 model and 12.58\% points for the GPT-4 model.
This suggests that the few-shot prompt strategy, which introduces data in the prompt, is effective. However, if the model's performance is near saturation (e.g., in the suicide risk task), no further improvement can be achieved. In contrast, in tasks where performance is initially poor, the few-shot strategy can lead to huge improvements.

The fine-tuning of LLMs achieved better performance in both tasks. For the suicide classification task, the fine-tuned GPT-3.5 model improved by 6.25\% points in F1-score compared to the zero-shot prompt. For the cognitive distortions task, the performance gains were even more significant, with GPT-3.5 improving by 61.03\% points in F1-score compared to the zero-shot approach. However, the performance difference between the fine-tuned GPT-3.5 and other fine-tuned models in the cognitive distortion task was not big, so we did not introduce fine-tuning comparison experiments of the other LLMs in the suicide risk classification task.
Despite the substantial improvements from fine-tuning, supervised learning still holds an advantage in both tasks. In the suicide classification task, BERT achieves an F1-score of 82.76\%, which is 4.31\% points higher than the best fine-tuned GPT-3.5. In the cognitive distortion task, BERT also achieves the best F1-score of 76.10\%, which is 4.31\% points higher than the best-performing LLM, Chinese-Alpaca-2-7B.
Although task-specific fine-tuning improves LLM performance, it still does not surpass supervised learning. Additionally, fine-tuning increases the difficulty and cost of using LLMs, diminishing their versatility and flexibility. The number of parameters in LLMs is generally much larger than in supervised learning models, making their training and use more expensive.

Overall, LLMs are effective for performing relatively simple tasks using only zero-shot prompt, but they need to be properly designed with prompt such as scene settings. For more difficult tasks, LLMs cannot perform adequately through prompt-only solution. The few-shot strategy, which introduces training data into the prompt, is effective but does not result in a significant performance improvement. Fine-tuning LLMs for specific tasks can achieve substantial improvements, but their performance still falls short of that achieved by supervised learning. Supervised learning remains the best choice for low-cost and high-performance in domain-specific tasks.

\section{Expert evaluation and feedback}\label{sec:result:Expert evaluation}
\begin{figure*}[!htbp]
\centering
\includegraphics[width=0.95\linewidth]{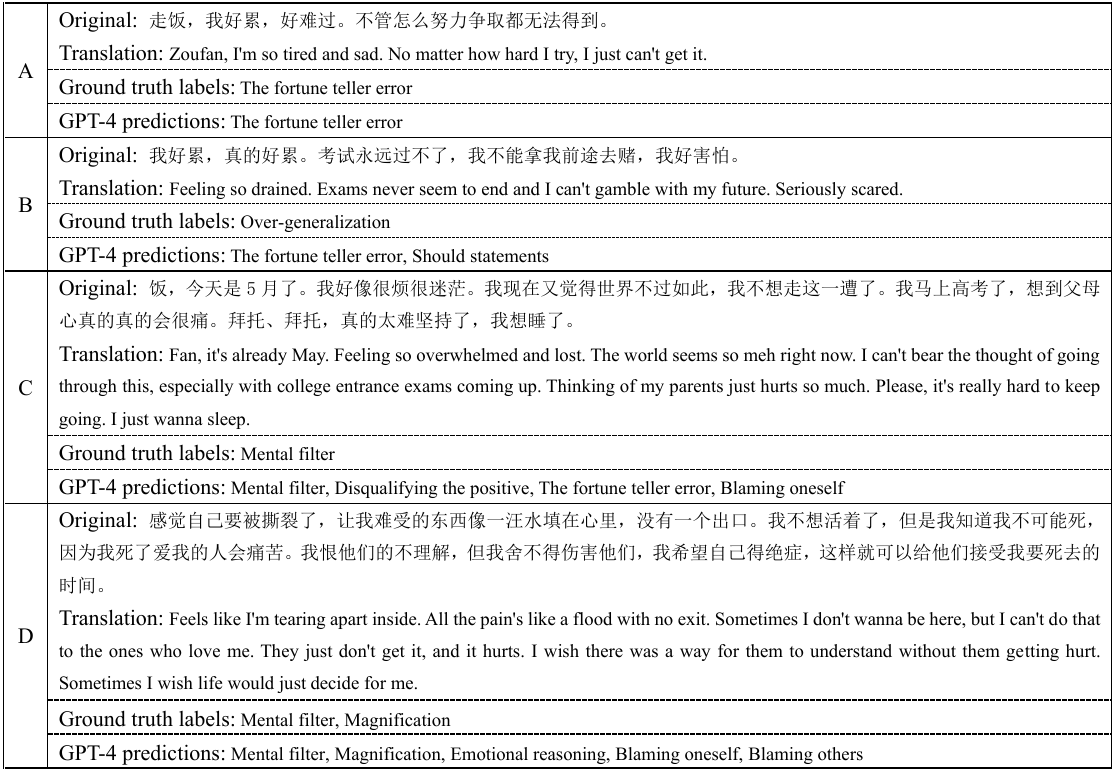}
\caption{Examples of expert annotated labels and GPT-4 predictions in cognitive distortion classification task.}
\label{fig:examples}
\end{figure*}

As mentioned above, using LLMs with prompts alone, without fine-tuning, is the best case scenario for their role. However, LLMs do not perform well under zero-shot and few-shot prompts. For this reason, we selected the output of GPT-4, which performs relatively well under the ``${\mathrm{ref}}_{2}$+basic'' strategy, for further analysis.
The examples can be seen in Figure~\ref{fig:examples}.

Overall, the model's judgments tend to be overly predictive and overly sensitive. When dealing with simple texts and scenarios, such as in Example A, the model can predict accurately. However, simple statements are not necessarily easier to identify; in fact, they may contain more ambiguous information, making them more difficult to evaluate. For example, in Example B, the model recognizes the negative prediction of the exam grade and the negative emotions, categorizing it as ``the fortune teller error'' and ``should statements.'' However, the model fails to recognize that the main feature of the sentence is the user's broad, generalized conclusion based on past test failures. This cognitive distortion aligns more with ``over-generalization'', as the writer infers that “the exams will never be passed” based on a limited number of failures.
These examples illustrate that while the model can make general judgments, it often fails to capture the central feature of the text, leading to incorrect predictions.

When dealing with longer texts, such as Examples C and D, the model is often able to find the correct category but can also over-predict. For example, in Example C, the model correctly identifies the category ``mental filter'' but also inaccurately predicts the other categories. These are incorrect because the sentence does not explicitly deny any positive experiences (disqualifying the positive), make unreasonable future predictions (the fortune teller error), or place blame on the self (blaming oneself).
Similarly, in Example D, the model correctly identifies ``mental filter'' and ``magnification'' but fails to capture that the main feature of the sentence is the user’s focus on negative details and exaggeration of the problem's severity. The model also incorrectly predicts three categories more, even though the sentence does not explicitly draw conclusions based on emotions (emotional reasoning), nor does it attribute blame exclusively to oneself (blaming oneself) or to others (blaming others).

These examples show that while models can make correct judgments in general direction, they are often overly sensitive to details, leading to overprediction and misclassification. The key to optimizing the model is to improve its ability to identify the central theme and main features of the text, thereby reducing unnecessary over-prediction.

\section{Conclusion} \label{sec:conclusion}
In this paper, we present two psychologically relevant datasets based on social media data: SocialCD-3K for cognitive distortion multi-label classification task and SOS-HL-1K for high/low suicide risk classification. We validate the performance of two supervised learning models and eight LLMs, including both Chinese and English models, on these tasks. Our experiments evaluate the performance of LLMs under different prompt designs. The results show that LLMs driven by prompts alone can perform well on the suicide risk classification task but fail in more complex tasks such as cognitive distortion recognition. While fine-tuning the LLMs can improve performance, it still falls short of supervised learning. Additionally, fine-tuning reduces the flexibility and ease of use of LLMs.
Our study provides new research data for the community and offers detailed experimental insights for those who intend to use LLMs in mental health domain.

\FloatBarrier
\bibliography{references.bib}
\bibliographystyle{IEEEtran}
\balance
\end{document}